# •A Novel Actuation Strategy for an Agile Bio-inspired FWAV Performing a Morphing-coupled Wingbeat Pattern

Ang Chen, Bifeng Song, Zhihe Wang, Dong Xue, Kang Liu

*Abstract*—**Flying vertebrates exhibit sophisticated wingbeat kinematics. Their specialized forelimbs allow for the wing morphing motion to couple with the flapping motion during their level flight. Previous flyable bionic platforms have successfully applied bio-inspired wing morphing but cannot yet be propelled by the morphing-coupled wingbeat pattern. Spurred by this, we develop a bio-inspired flapping-wing aerial vehicle (FWAV) entitled RoboFalcon, which is equipped with a novel mechanism to drive the bat-style morphing wings, performs a morphing-coupled wingbeat pattern, and overall manages an appealing flight. The novel mechanism of RoboFalcon allows coupling the morphing and flapping during level flight and decoupling these when maneuvering is required, producing a bilateral asymmetric downstroke affording high rolling agility. The bat-style morphing wing is designed with a tilted mounting angle around the radius at the wrist joint to mimic the wrist supination and pronation effect of flying vertebrates' forelimbs. The agility of RoboFalcon is assessed through several rolling maneuver flight tests, and we demonstrate its well-performing agility capability compared to flying creatures and current flapping-wing platforms. Wind tunnel tests indicate that the roll moment of the asymmetric downstroke is correlated with the flapping frequency, and the wrist mounting angle can be used for tuning the angle of attack and lift-thrust configuration of the equilibrium flight state. We believe that this work yields a well-performing bionic platform and provides a new actuation strategy for the morphing-coupled flapping flight.**

## I. INTRODUCTION

BIRDS and bats can change their wing planform, not only during gliding but also in downstroke and upstroke phases during the level flapping flight. Research suggests that birds and bats reduce their wingspan and area during the upstroke by rapidly withdrawing the wings towards the body and increase these parameters during downstroke [Fig. 1(a)] [1], [2]. This wingbeat kinematics presumably reduces drag and enhances aerodynamic efficiency during level flight [3]-[6].

Despite the wing's structure details are different, flying vertebrates (including pterosaurs) share several standard features regarding their forelimbs [Fig. 1(b)] [7], [8]. Specifically, they all have an arm wing that comprises the humerus and forearm (radius and ulna) and a hand wing spanned by manus and other structures (feathers or membranes). To maintain rigidity and lightness, the wing articulation constraints the forelimb motion primarily to those required during flight [9]-[13]. Since bats can bend and stretch their wing joints, bird wings have fewer joints and less independent controllable degrees of freedom (DOFs) than bat wings. Due to their wing skeletal system, birds can automatically couple extension and flexion of their wing joints, making elbow and hand extend together and simplifying feathered wing morphing [13], [14]. In analogy, both these flapping flyers can tuck and extend their wings on a unilateral side, which morphs the wing shape and results in the span and area change [2], [15].

Span and area reduction of upstroke vary for different species and flight velocities. Birds sweep their hand-wing backward during upstroke while their arm-wing remains partially extended, benefiting from independent overlapping feathers [15]-[18]. In contrast, microbats reduce their arm-wing span while keeping the hand-wing membrane relatively stretched, and megabats bend fingers to retract the hand-wing [19], [23]. It is commonly believed that birds and bats exhibit different upstroke kinematics in level flight [18], but researchers have also indicated that some bat species apply a hand-wing back-sweep which is similar to that of birds during upstroke when they fly faster [24]. Both birds and bats exhibit highly specialized wrist osteology, making the hand-wing execute supination during the upstroke and prevent hyperpronating during downstroke by exploiting an interlocking mechanism within their wrist osteology [25]-[27]. Flying vertebrates also use wing morphing and wingbeat kinematics to enhance their flight performance and agility. For instance, an extended wing is advantageous for slow gliding and turning, while a swept wing is suitable for fast gliding [28]. Since bat wings are proportionally heavier than bird wings [29], bats change wing inertia to execute falling and landing maneuvers, while birds rely predominantly on redirecting aerodynamic forces for low-speed maneuvers [30]. To initiate a rolling maneuver, birds generate bilateral asymmetries in the wing morphing and wing

This work was supported by National Natural Science Foundation of China, Grants No.11902103 and 11572255. *(Corresponding author: Dong Xue.)*

The authors are with the School of Aeronautics, Northwestern Polytechnical University, Xi'an, Shaanxi 710072, China (e-mail: 15754602650@163.com; sbf@nwpu.edu.cn; cfrpg@mail.nwpu.edu.cn; xuedong@nwpu.edu.cn; kangliu@mail.nwpu.edu.cn).

trajectory to produce roll moment [30], [31]. This asymmetric morphing coupled with wingbeat gives birds the high agility to perform extreme maneuvers for foraging and to escape from predators. Both birds and bats can achieve their body reorientation within a few wing-beats [29], [30], e.g., in the most extreme maneuver, a hummingbird can do a 180° yaw turn even within four wingbeats [32].

Inspired by the wing anatomical mechanism and versatile flight performance of birds and bats, many bionic robots and aircrafts have been developed to mimic the wing morphing motion of those airborne creatures. Some cases have successfully improved flight agility and maneuverability by using morphable articulated wings with membranous or feathered surfaces [15], [33]-[38]. Unlike their natural counterparts, most of those works are validated on conventional platforms, which used propeller for thrust modulation or elevator/rudder for attitude control [15], [33]-[35]. While some flapping-wing designs have also been used, these platforms actuate the wing morphing independently, which is not associated with wing flapping [36]-[38].

However, for flying vertebrates, the change of wingspan and area is coupled with wingbeat, because as already discussed, they use different wingspan and area for up- and down-strokes. Current bio-inspired platform designs lack this ability for a wing-extended downstroke and a wing-tucked upstroke with bio-inspired morphing wings, while investigations of this wingbeat pattern for bionic robots or aircrafts are usually limited to wind tunnel tests and simulations [39]-[42].

Nevertheless, applying these wingbeat kinematics to a flyable platform imposes a significant engineering challenge for the mechanism and actuator design. Indeed, the wing extended and tucked motion must be actuated at the same frequency as the flapping motion, which is required for relatively high power and high-speed actuators. However, bio-inspired flapping wings need to be lightweight to reduce the moment of inertia for high-speed flapping, which brings restrictions to equip proper actuators for wing morphing coupled with flapping.

Here, we develop a flyable bio-inspired flapping-wing aerial vehicle (FWAV), the RoboFalcon, equipped with a specially designed novel mechanism that achieves the wingbeat kinematics of birds during level flight with a pair of bat-style membranous morphing wings [Fig. 1(c)]. Specifically, this platform extends its wings during the downstroke phase to generate lift and thrust and tucks the wings in the upstroke phase. The proposed mechanism couples wing flapping motion with wing morphing motion and allows the servo to actuate the wing morphing motion during downstroke by briefly decoupling morphing from flapping when maneuvers are required. The wrist joint of the morphing wing is designed with a pitch-up mounting angle to imitate the wrist supination of the bird and bat wing for tuning passive twist of the wing during the downstroke. Our design results in a well bionic-performing FWAV platform with high rolling agility, as RoboFalcon manages 90 degrees rolling maneuver within 2.5 wingbeat cycles. We validate RoboFalcon's rolling agility by performing several flight experiments, where each experiment rolling maneuver is initiated with a one-shot bilateral asymmetric downstroke flapping. We also analyze the flight characteristics of four different wrist mounting angles and measure the roll moment of the bilateral asymmetric downstroke via wind tunnel tests. The results indicate that the wrist mounting angle could affect the angle-of-attack and lift/thrust magnitude of the equilibrium flight state, suggesting that birds' and bats' wrist supination/pronation may play a similar role in level flight. Results also highlight that the roll moment generated by the asymmetric wingbeat pattern is correlated with flapping frequency, affording high roll control efficiency. Overall, we believe that this work provides a new actuation strategy for morphing-coupled wingbeat patterns taking into account lateral control, and suggests a well-flying bionic platform for investigating the flapping flight for flying vertebrates.

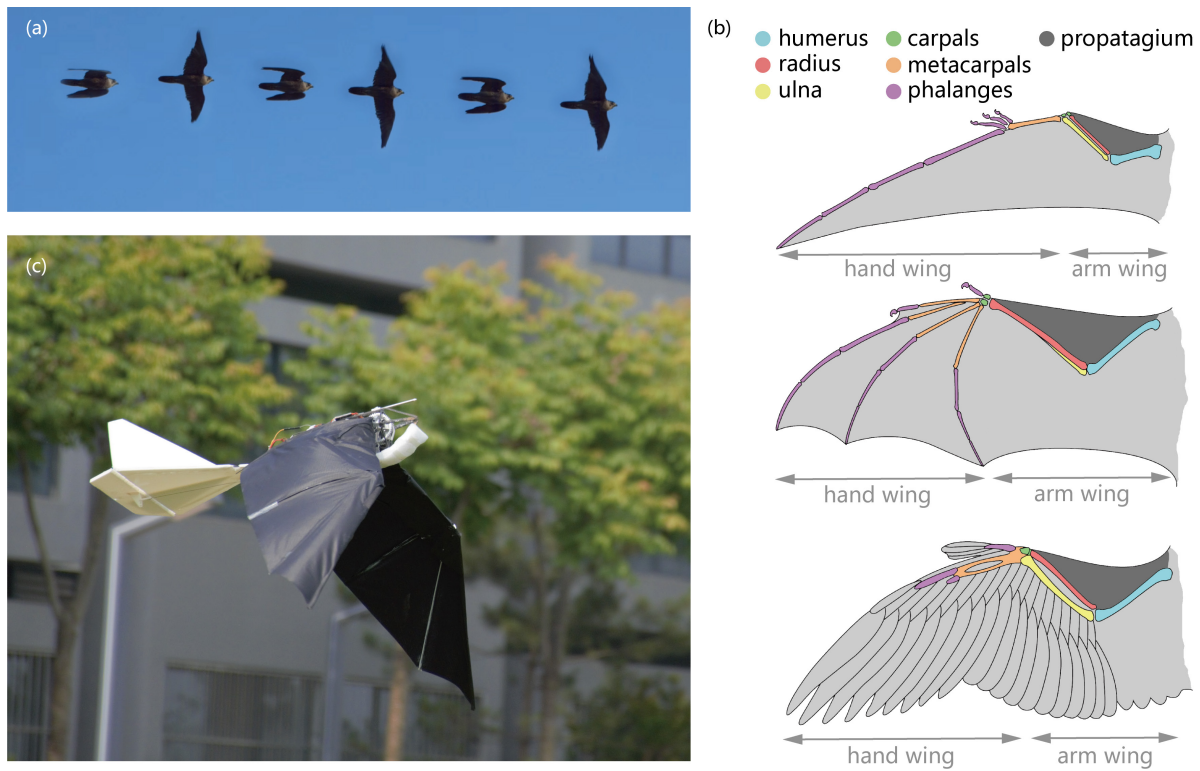

Fig. 1. The wingbeat pattern of avian and the forelimb anatomy of flying vertebrates inspired the design of a bionic FWAV. (a) Snapshot of a peregrine falcon during level flight. The wing is tucked in upstroke and extended in the downstroke. (b) Wing anatomy of pterosaur, bat, and bird, exhibiting analogous three-link structures. (c) The RoboFalcon FWAV in flight, as inspired by flying vertebrates, applying a novel actuation strategy for achieving the wingbeat pattern of the bird in (a).

## II. PLATFORM DESIGN

To reproduce the actual flight effect of the flying creature and validate the flight capability of the mechanically synthesized wingbeat kinematics, the design of the flight platform is better referenced to medium- or large-sized birds which exploit this flapping pattern during level flight. Our RoboFalcon FWAV platform is generally based on the morphological parameters of the peregrine falcon (*Falco peregrinus*), which is designed with a maximum wingspan of 1.2m and a total flight weight of 600g [Fig. 2(a)]. RoboFalcon is equipped with a conventional tail made from foam board that is strengthened utilizing carbon fiber rods, with the latter being embedded in the leading edge of the tail [Fig. 2(b)]. The tail involves only an elevator providing flight pitch control. The body frame is made from 2mm carbon fiber composite boards cut into specifically shaped pieces utilizing CNC machinery to form a 3D puzzle structure

[Fig. 2(c)]. The flight controller placed in the body frame is an autopilot (PixHawk mini) equipped with an integrated inertial measurement unit (IMU). Other accessories include an RC receiver for control command input, a power management module for power input measuring, and a pitot tube for airspeed sensing [Fig. 2(d)]. With this configuration, it is possible to record on a memory card the attitude angle and angular velocity in the 3-axis, the airspeed, and the power consumption during the flight. RoboFalcon is powered by a Li-Po battery (3s 850mAh) and equipped with a brushless DC motor (SunnySky X2212 KV980) to drive the flapping mechanism [Fig. 2(d)].

## III. BIO-INSPIRED MORPHING WING

To simplify the structure and allow for a lightweight design, we choose for the bio-inspired morphing wing design the bat-style membranous wing structure over the bird-style feathered (Fig. 3). Despite naturally bird wings are proportionally lighter than bat wings [15], [29], the artificial version of bat-style morphing wings affords a lighter design than bird-style wings because artificial feathers made from carbon or glass fiber are heavier and less robust than real feathers [34].

Our bat-style morphing wing's inner section (arm wing) comprises two main spars hinged to each other, representing the humerus and radius [Fig. 3(a)]. These spars are constrained by several linkage systems ensuring the extension and flexion of the arm that are determined by the linear distance between the shoulder and the proximal end of lever 1 [Fig. 3(a), marked with red two-way arrows]. Such design results in a one-DOF sophisticated multi-link mechanism analogous to the wing designs in [37], [38]. These spars are made by the same technique as the body frame, and their junction forms a robust carbon fiber revolute joint, the elbow joint [Fig. 3, (b) and (c)]. The wing's outer section (hand wing) is spanned by three fingers, with the outermost being the strongest one forming the leading edge and the other two supporting the wing surface. The fingers are designed to curve downward to help create a positive curvature airfoil and generate as much as possible lift [Fig. 3(b)].

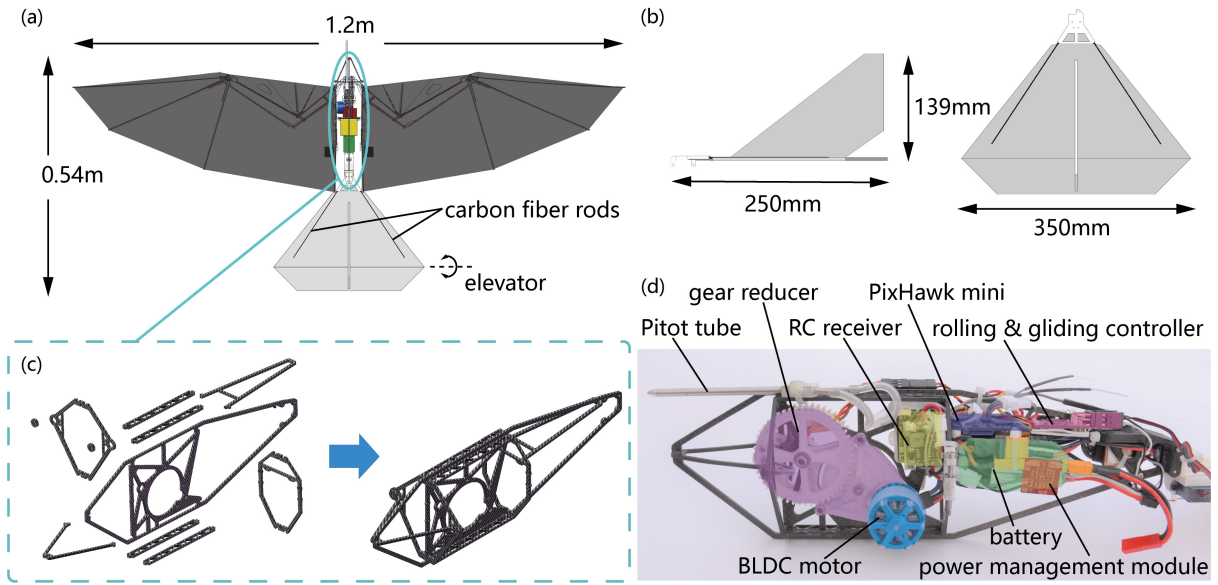

Fig. 2. RoboFalcon platform design and subsystem configuration. (a) Top view of RoboFalcon with the wings fully extended at level position. The wingspan is designed concerning the maximum wingspan of the peregrine falcon [60], while the total weight is slightly lower than average for this species, which enables a lower wing loading. (b) The tail made from foam board has a conventional elevator for pitch control, and a vertical stabilizer achieves the yaw stability without a rudder. (c) The CNC-cut carbon fiber parts form the RoboFalcon's body frame. (d) Avionics, sensors, and power transmission mechanism mounted on the body frame.

The anterior ends of these fingers are hinged to the distal end of the radial spar. More specifically, the finger on the leading edge is connected to the radial spar with a revolute joint [wrist joint, Fig. 3(d)], while the other two are connected with ball-and-socket joints. These two fingers are entirely constrained by the link rod [rod 3, Fig. 3(a)] that is posterior to the radial spar with pin-slot joints [Fig. 3(d)], allowing the fingers to be actuated by the extension and flexion motion of the arm wing.

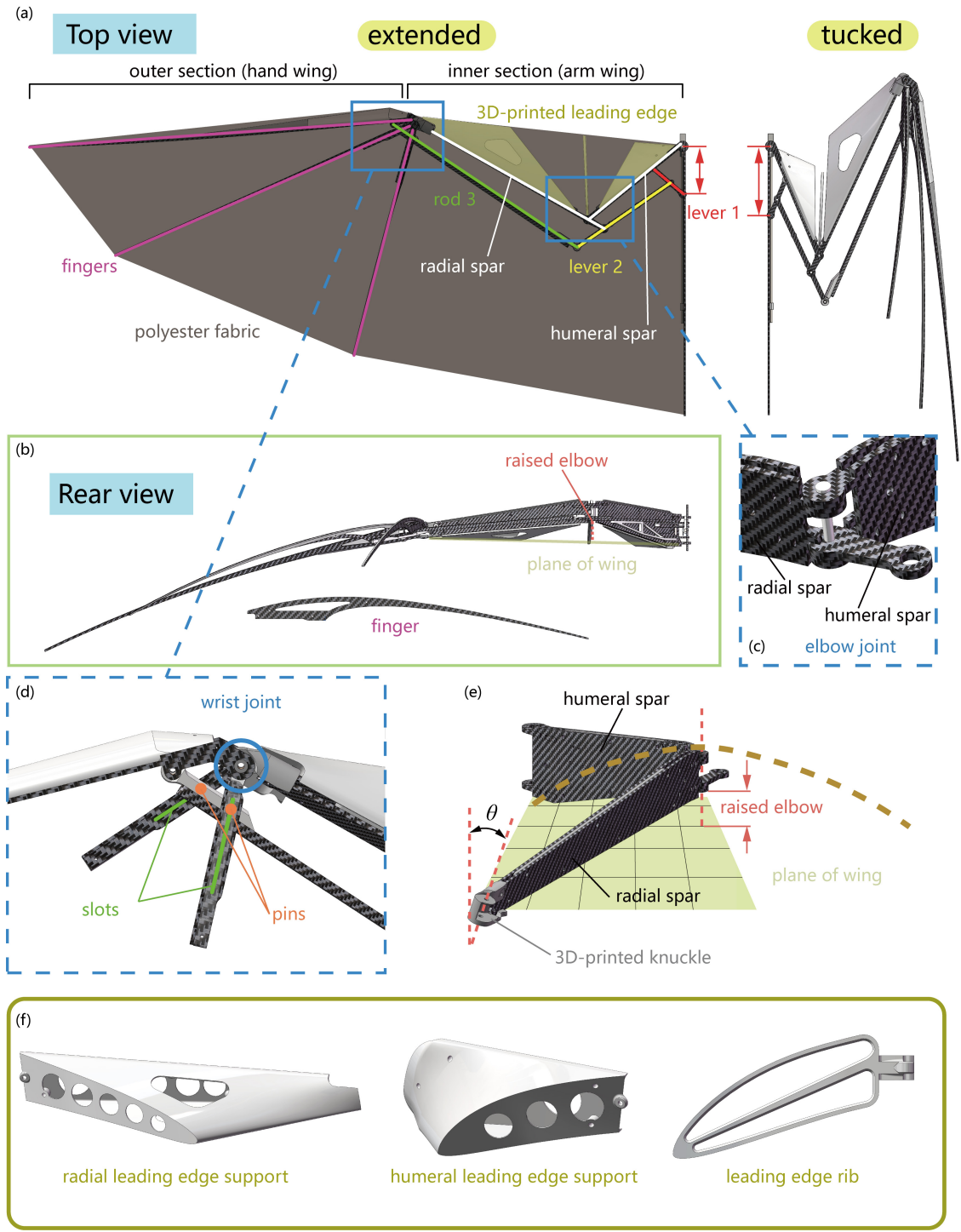

Fig. 3. Bio-inspired morphing wing with the bat-style layout and other bionic details. (a) Top view of a fully extended wing (left) and tucked skeleton (right). The morphing motion is driven by lever 1, denoted in red. (b) A rear view of a fully extended wing skeleton along with an arched finger is shown below. (c) Details of the elbow joint hinge. (d) Enlarged top view of the wrist joint. Two of the fingers are fully actuated with pin-slot joints in morphing motion. (e) The elbow joint is designed above the wing plane that helps to increase the camber of the membrane-covered arm wing, which is not easy to manage an accurate aerodynamic profile. The 3D printed knuckle is tilted by a certain angle θ around the radial spar at the distal end, which allows the rotation axis of the wrist hinge to be tilted relative to the wing plane and helps the out-of-plane motion of the hand wing. (f) The 3D-printed leading-edge support parts help maintain the profile of the triangular area, in (a), spanned by the humeral and radial spar, which plays the role of propatagium.

Given that bats and birds can both change their wing camber, some researchers have indicated that these flying creatures adjust their proximal wing camber by a flexible wing membrane called the propatagium [Fig. 1(b)] combined with a change in elbow joint posture [43]-[45]. Inspired by how bats and birds change their wing camber, we purposely place the elbow joint at the highest point of the wing surface to help achieve a high inner section camber [Fig. 3, (b) and (e)]. The wrist joint [Fig. 3(d)] has only one degree of freedom, prohibiting wrist supination and pronation. Given that such supination and pronation exist in birds and bats, we circumvent this problem by an adequately adjusted wrist mounting angle, as mentioned in the later text [46], [47].

We construct the wrist joint with a pitching-up mounting angle by attaching a 3D printed knuckle to the distal of the radial spar [Fig. 3(e)]. The 3D-printed knuckle tilts at a certain angle θ around the radial spar, and thus the rotation axis of the wrist joint is not perpendicular to the wing plane. This arrangement results in a pitching-up hand-wing that can partially offset the passive wrist pronation (hyperpronating) due to the aerodynamic load on the wing during the fully extended downstroke. Since the wing skeleton is retracted during the upstroke, the pitching-up wrist joint causes the hand-wing to bend downward even more, achieving the same supination and flexion movement that birds and bats do in upstrokes [25], [26].

The elastic membranes have been successfully applied to bat-inspired wings [37], [38], however, during a constantly morphing motion, the elastic damping of these materials imposes extra energy consumption. Instead, bats actively change their membrane stiffness to keep the wing surface flat during the morphing process [48]. Considering the difficulty of fabricating the flexible membrane materials that can actively and rapidly change the stiffness over a large scale, we cover the wing skeleton with the unstretchable polyester fabric as wing membrane to avoid burdening the morphing motion actuator [Fig. 3(a)]. This arrangement may cause unrestrained collapse and wrinkling of the wing surfaces during skeleton retracting and ensure a solid and stable surface when the wing is fully extended. This collapse and wrinkling do not highly impact the flight (it may cause some extra drag, but RoboFalcon is still flyable), as discussed later. Two lightweight 3D printed parts are mounted in front of the humeral and the radial spar to support the front wing membrane, representing the propatagium, to form a relatively rigid leading edge [Fig. 3(a), 3(f)]. The 3D-printed parts are designed to preserve the aerodynamic shape of the leading edge and avoid interference when the skeleton is retracting. Additionally, a 3D printed rib [Fig. 3(f), rightmost] hinged to the elbow joint helps support the wing membrane in the gap between the two 3D printed parts [Fig. 3(a)].

As a result, the raised elbow and twisted wrist help create a highly cambered membranous morphing wing with a rigid but deformable leading edge. The wing can be passively twisted under aerodynamic loads to generate lift and thrust during the downstroke. The two ends of the link rod 3 [Fig. 3(a)] are equipped with ball-and-socket joints, which allows the passive torsion of the outer section skeleton to depend primarily on the radial spar. Hence, we can fine-tune the passive torsion of the hand wing by adjusting the mounting angle θ of the 3D printed knuckle of the wrist joint for optimal aerodynamic forces (as thoroughly discussed later). Since the torsional stiffness of the radial spar remains the same, the larger θ, the weaker the passive torsion during the downstroke.

## IV. Flapping & Morphing Coupling Mechanism

### A. Conical Rocker Mechanism

To achieve the coupling linkage of wing flapping and wing morphing, we design a mechanism (Fig. 4) that allows flapping and morphing to be actuated at the same frequency. The phase of the flapping motion can be 90 degrees ahead of the periodic morphing motion, which enables the wing to extend during the downstroke and to tuck during the upstroke.

This mechanism is named the Conical Rocker Mechanism (CRM). The mechanism uses a rocker arm [denoted by blue in Fig. 4(a)] to convert the rotation motion of a gear [denoted by orange in Fig. 4(a)] into the flapping motion of the humeral spar. A shaft (shaft C) eccentrically hinged to the gear with a ball-and-socket joint act as an oblique crank [denoted by green in Fig. 4(a)]. The shaft C is confined in a shaft hole of the rocker arm and is only allowed to rotate and slide with respect to the rocker arm [Fig. 4(b)]. The rocker arm and the humeral spar are connected by a shorter shaft, while another longer shaft, oriented at 90° to the shorter one, is restrained by the body frame. The cross shaft [denoted by red in Fig. 4, (a) and (c)] formed by these two shafts restrains the motion of the rocker and the humeral spar. As the gear rotates, shaft C drives the rocker arm to sway in a conical trajectory [Fig. 4(d)], while the center of the cross shaft is located right at the apex of the cone. As actuated by the rocker, the cross shaft oscillates around the longer shaft causing the humeral spar to rotate up and down, affording wing flapping.

This mechanism has many advantages in adjusting flapping-wing parameters such as amplitude, the tilt angle of the stroke plane, and dihedral angle. For instance, the flapping amplitude is defined by the apex angle of the cone, which is described by

$$\Phi = 2\arctan\frac{R}{H} \qquad (1)$$

where $\Phi$ is the flapping amplitude angle, $R$ the radius of the crank (distance between the rotation axis of the gear and the ball-and-socket joint of the shaft C), and $H$ the height from the rotation plane of the crank to the center point of the cross shaft. Since shaft C is allowed to slide in the rocker arm, we can change $H$ by moving the center point of the cross shaft laterally, thus changing the flapping amplitude. Similarly, because shaft C is hinged to the gear by a ball-and-socket joint, the cross shaft is allowed to move upward or downward to force the rocker arm to sway in an oblique conical trajectory, thus changing the dihedral angle. Moreover, tilting the longer shaft of the cross shaft around the lateral or the vertical axis causes a change in the tilt angle of the stroke plane or the wing to sweep back or forth. By adjusting the position of the cross shaft relative to the body frame with appropriate actuation methods, we can adjust these parameters dynamically without interfering with flapping.

These advantages of CRM could help develop the insect- or hummingbird-inspired FWAV. However, it should be noted that in this article, we focus more on CRM's feature of achieving the coupling motion of flapping and morphing.

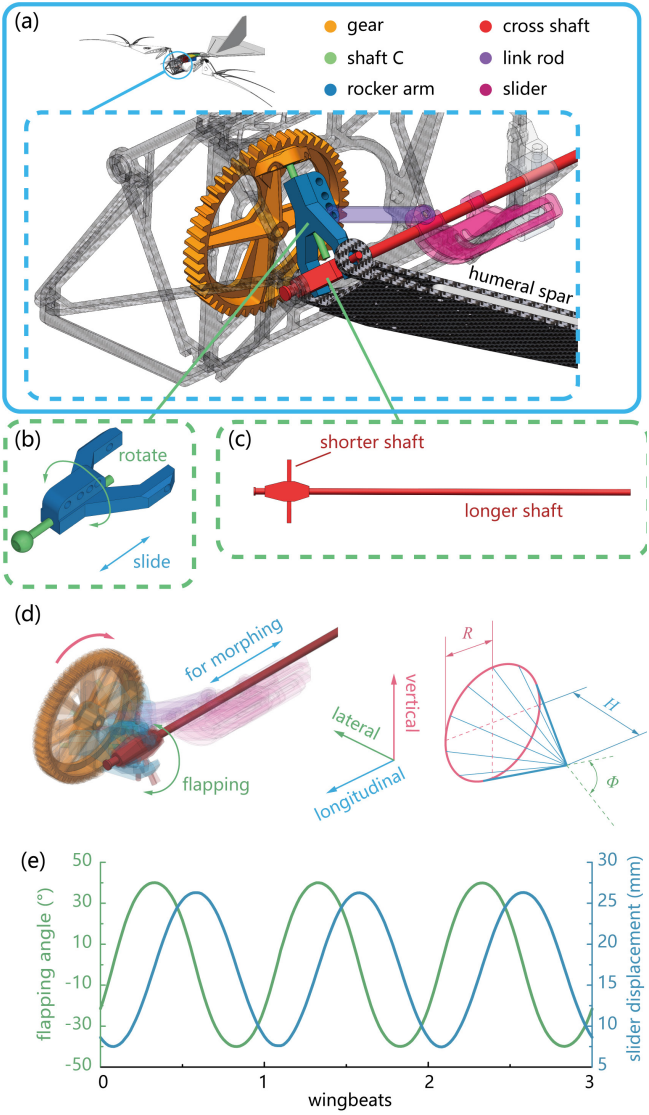

Fig. 4. Wingbeat actuation mechanism for flapping and morphing coupling. (a) The Conical Rocker Mechanism (CRM) is mounted in the body frame, with its components marked in different colors. (b) The mechanical constraints of the rock arm and shaft C. (c) The cross shaft is the critical part of CRM. The longer shaft acts as the flapping axis of the wing, while the shorter as the rotation axis of the shoulder when performing wing morphing. (d) The action effect of CRM and the conical trajectory of the rocker arm. (e) The flapping motion is phase lead the morphing actuation motion by 90°.

To derive the conical swing of the rocker arm into a linear reciprocating motion, a link rod [denoted by purple in Fig. 4(a)] is hinged between the rocker arm and a slider [denoted by magenta in Fig. 4(a)] on the longer shaft. With the slider's reciprocating motion, we can easily actuate the wing skeleton to extend and tuck periodically by hinging lever 1 [in Fig. 3(a)] to the slider. Thus, the wing morphing motion will be coupled with the wing flapping motion with 90 degrees phase lag [Fig. 4(e)]. However, this arrangement will cause the wing to extend fully only when it flaps down to the level position. Since morphing and flapping are coupled, wing morphing cannot be used independently for attitude and maneuvering control, as birds and bats do.

## B. Morphing decoupler

To ensure the wing is fully extended during the downstroke and the wing morphing motion can be controlled independently while maneuvers are required, we introduce a morphing decoupler to meet these requirements [Fig. 5]. The decoupler comprises three sliders: motor input slider (MIS), servo input slider (SIS), and output slider (OS) [Fig. 5, (a), (b), (c)]. All these sliders are constrained to allow only linear sliding along the longer shaft of the cross shaft. The MIS linked with the rocker arm is powered by the BLDC motor and moves reciprocally. The SIS is actuated (or locked) by a servo mounted at the posterior end of the longer shaft through a link rod [Fig. 5(a)]. MIS and SIS are processed by a five-axis CNC machine, and both have sliding arms located outside the longer shaft with specially shaped slots on them [Fig. 5(c)]. The sliding arm of MIS holds the sliding arm of SIS within it, sharing a pin located in their slots [Fig. 5(b)]. The pin is linked to the OS, and the OS is hinged to lever 1 in Fig. 3(a). This design constrains the OS sliding by MIS and SIS, which means that the wing morphing motion can be actuated by both the BLDC motor and servo. Specifically, the relationship between the motion of these three sliders is divided into two cases. In the first one, when the MIS slides in a relatively anterior position, the pin is simultaneously in the linear slot of MIS and the curved slot of SIS. Hence, the pin can slide with respect to the MIS while being limited by the SIS [shown on the upper side in Fig. 5(d)].

Regarding the second case, when the MIS slides posteriorly, the pin is in the curved slot of MIS and the linear slot of SIS, and thus, the pin moves together with the MIS and can slide with respect to the SIS [shown downside in Fig. 5(d)]. In this way, the OS, which outputs power to the wing, will be driven by the motor or locked by the servo in due time with the motion of the pin. If the gear turns in a direction as in Fig. 5(a) (indicated by yellow arrow), the wing would be fully extended during the downstroke because the wing is decoupled from the MIS that is moving anteriorly and locked in the extended state by the SIS. When the MIS moves at the rear with the gear rotation, which occurs during the upstroke, the wing is again coupled to the MIS and performs the morphing motion actuated by the motor. Equipped with the morphing decoupler, the CRM can drive the wing to create the wingtip and wrist trajectory as shown in Fig. 5(e), which is significantly similar to the bird trajectory of previous studies [49].

Furthermore, since the SIS locks the wing in the extended state during the downstroke, the morphing motion of the wing can be controlled by the rear-mounted servo [Fig. 5(f)]. Therefore, the left and right wings can be actuated independently to cause a bilateral asymmetric change in the wingspan and area during the downstroke phase. In this way, the servos on both sides can ignore the time sequences of wings flapping and move their respective SISs to the appropriate positions to ensure that each wing is or is not fully extended during each downstroke. This design greatly eases the task of the wing morphing actuator and reduces the difficulty of lateral attitude control for the FWAV with a complex wingbeat pattern. The servos do not need to run at the same frequency as the wing flapping, affording lightweight and low-powered servos. For example, to perform a left rolling maneuver, we can input a command to make the left servo pull the left SIS backward to limit the maximum wingspan on that side and then allow the wing to flap for a few cycles. During this process, the wings on the two sides flap downward asymmetrically to generate a differential lift and then tuck and flap upward symmetrically driven by the CRM (see supplementary video). This feature allows the RoboFalcon to be controlled manually without an autopilot.

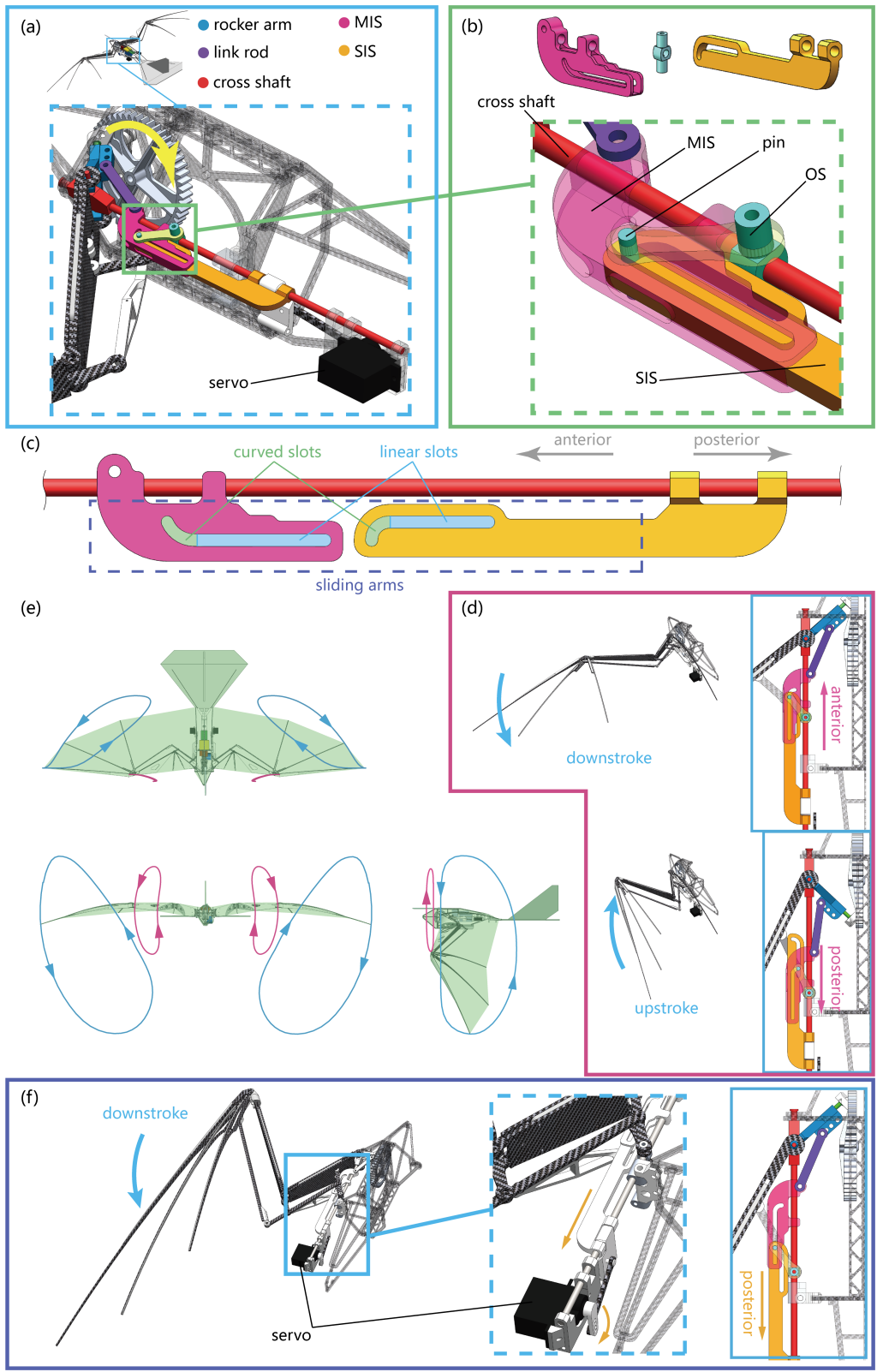

Fig. 5. Wing-morphing decoupling mechanism for roll control. (a) The morphing decoupler is mounted on the longer shaft of the cross shaft, with each part denoted in different colors. (b) Enlarged view of the morphing decoupler. For the sake of clarity, the trimetric views of these three sliders (MIS, OS, SIS) are shown above to exhibit their 3D geometry. (c) The top view of MIS and SIS aligned on the longer shaft. Both sliders have hockey-stick-shaped slots on their sliding arms to guide the pin's motion in (b). (d) With the morphing decoupler, the wing skeleton is fully extended during downstroke by decoupling from the MIS (shown above) and tucked during upstroke by coupling with MIS again (shown below). (e) The wingtips and wrists trajectory of the wingbeat kinematics driven by CRM with the application of morphing decoupler. (f) The servo actuates the wing to tuck in during downstroke because of MIS decoupling (see supplementary video for more details).

## V. Rolling maneuver & flight test

Although the morphing decoupler allows the servo to actuate at any moment during the flapping flight, we need a definite and controllable bilateral asymmetric downstroke to realize how the morphing-coupled wing-flapping pattern utilizes the asymmetric downstroke to perform a roll maneuver. For this purpose, we equip the RoboFalcon with a simple Hall-sensor-based closed-loop control system to detect the flapping sequences of the wings [Fig. 6, (a) and (b)]. The Hall switch, mounted at the body frame [Fig. 6(a)], is triggered by a magnet embedded in the gear when the motor drives the wings flapping down to the level position. As the gear rotates, the rolling and gliding controller [Fig. 6(b)], based on STM32 F103 (see Section VII), determines the current flapping cycle period by measuring the time interval between two recent sequential triggers of the Hall switch. Once the roll command from the autopilot is received, the controller can manage the action time of the servo with the known wingbeat cycle period to ensure that the unilateral wingspan is limited only during the next single downstroke phase (see Section VII). Thus, we obtain a single and stable bilateral asymmetric downstroke among a sequence of standard symmetric flapping. Furthermore, we program the rolling and gliding controller to stop the motor only when the Hall switch is triggered. By adding a glide-lock to stop the gear reversal [Fig. 6(A)], the wing will be locked at the level position, ensuring for RoboFalcon the gliding ability and the capability to use the bilateral asymmetric wing morphing during gliding to achieve the banking turn of [33], [34].

To demonstrate RoboFalcon's roll control and high agility, we perform several rolling maneuver experiments to measure pitch, yaw, and roll angles (airspeed, power consumption, linear acceleration, and angular velocity were also recorded). Each maneuver is initiated with a single cycle bilateral asymmetric downstroke at an 8m/s level flight (with tethered string protection, see Section VII). Then the vehicle preserves the typical symmetric flapping during the open-loop rolling dynamic response [Fig. 6, (c) and (d)]. Each rolling maneuver utilizes the same servo value to ensure that each asymmetric downstroke has the same area difference between the left- and right-wing. An aircraft's agility is commonly defined as the rapidity of change in speed and direction, typically measured through the agility metric that includes angular acceleration and the time it takes to achieve a desired rotation angle change [50], [51]. In some studies, the time normalized by the wingbeat-cycle period is often used to quantify the flapping-wing flight maneuvers [52]. To compare the agility differences between various bio-inspired flapping-wing aerial vehicles and flying creatures, we use the agility metric measured as the time normalized by wingbeat-cycle that the flapping wing flyers require for a 90-degree attitude change. In Fig. 6(e), we compare the rolling agility of RoboFalcon against other flapping wing vehicles and flying creatures.

Our results indicate that the asymmetric downstroke based on the morphing-coupled flapping notably improves the rolling agility of the flapping wing flight [Fig. 6(d)]. During the asymmetric downstroke, the vehicle starts to roll left, and the roll angle decreases after maneuver initiation, indicating a lack of lateral stability for the RoboFalcon. This lack of lateral stability is partially the reason for its high rolling agility, with the roll rate being higher in each downstroke than in the upstroke, which is probably due to the rightward yaw angle of the vehicle during the rolling maneuver. The adverse roll-yaw coupling due to asymmetric wing morphing was also found in a previous study on a propeller-powered platform in a gliding state [34]. Nevertheless, for the asymmetric morphing coupled with flapping, the mechanism of adverse yaw generation could be more complex, which is not within the scope of this paper. The RoboFalcon achieves a 90-degree rolling maneuver within 2.5 wingbeats, which is an appealing performance compared to previous bio-inspired flapping-wing robots [32], [37], [52], and at the same level compared to some flying vertebrates [29] [Fig. 6(e)].

In our untethered outdoor flight test, RoboFalcon flies quite well, managing a climb, banking turn, and rapid rolling maneuver [Fig. 6(f) and supplementary video]. This agile vehicle applies a simple PID controller of the autopilot for pitch and roll control and can be remotely operated by a human pilot. The power consumption for an 8m/s level flight is about 50W, enabling a flight endurance of 11 min.

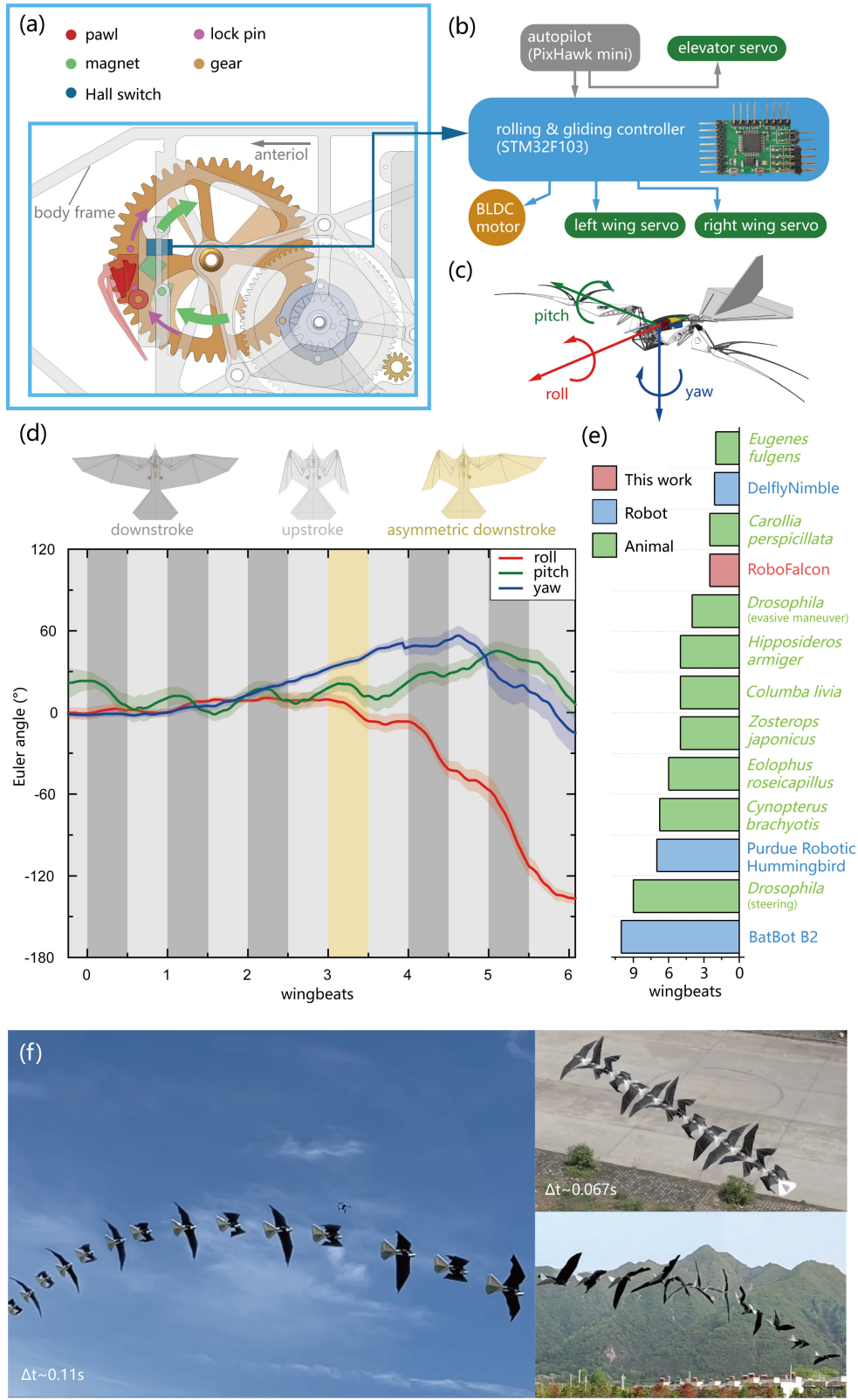

Fig. 6. The rolling maneuver for agility assessment and the flight test. (a) The glide-lock mechanism. The Hall switch (denoted in blue) is for wing position checking. The pawl (denoted in red) stops the gear from reversing when the wing flaps down to the level position. (b) The block diagram of the rolling and gliding control system. The autopilot controls the elevator servo directly, while the control of the motor on the wing servos is via the rolling & gliding controller. (c) The reference frame of the Euler angle. (d) The time evolution of the Euler angles during the rolling maneuver. The maneuver is initiated with a one-shot bilateral asymmetric downstroke (denoted in yellow). Symmetric wing-extended downstrokes are indicated in dark gray and wing-tucked upstrokes in light gray. (e) Comparison of the wingbeat number for performing a 90-degree orientation change for different flapping flyers. The data are acquired from [29], [32], [61]-[67] for the animals and [32], [37], [52] for the robots. (f) RoboFalcon performing a banking turn (shown left), a climb (shown upper left), and a rapid rolling maneuver (shown lower left) during flight tests.

## VI. WIND TUNNEL TEST

To assess the aerodynamic effect of the wrist mounting angle (wrist supination) and the characteristics of the bilateral asymmetric wingbeat kinematics, we measure the aerodynamic forces with a 6-axis load cell under various wing configurations in a low-speed wind tunnel (see Section VII). Firstly, we test the effects of four different wrist mounting angles θ (10°, 15°, 20°, and 25°) on the flight characteristics of a steady level flight state. The tests involve measuring lift and net thrust with various flapping frequencies at an airspeed of 8m/s and angles of attack ranging from 2 to 12 degrees placed below stall.

Significant cyclical aerodynamic effects characterize the flapping wing, and thus analyzing the steady flight state is not trivial. From the cycle-averaged perspective, the flapping wing aerodynamics can be analyzed regarding the method of the fixed wing [53] (see Section VII). The cycle-averaged lift and cycle-averaged net thrust of various flapping frequencies versus angle of attack are illustrated in Fig. 7(a).

During the trimed steady level flight, the average aerodynamic forces acting on the vehicle are at equilibrium, i.e., the cycle-averaged net thrust should be zero, and the cycle-

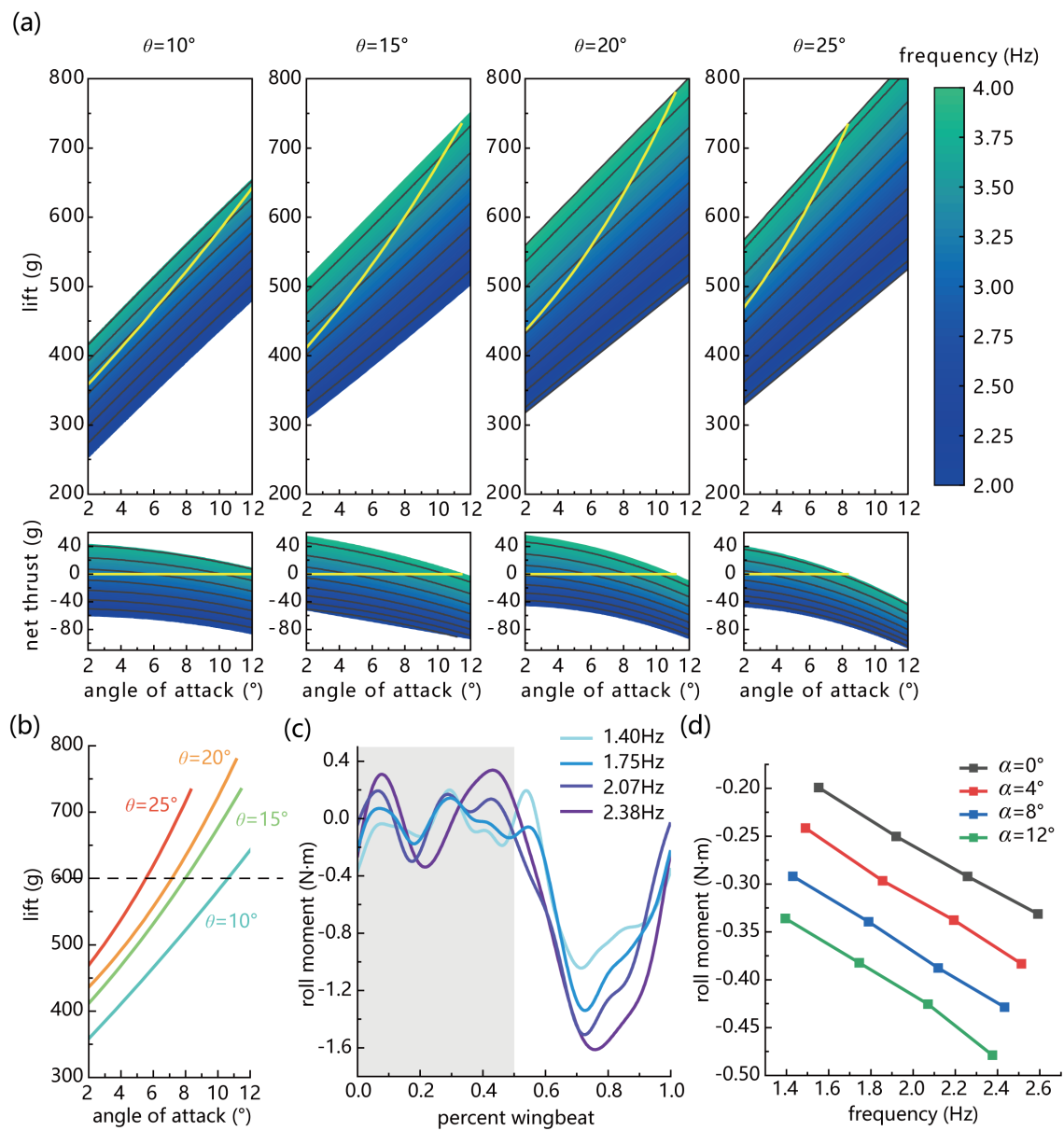

Fig. 7. The effects of wrist mounting angle and the roll moment of the asymmetric downstroke. (a) The cycle-averaged lift and net thrust of four different wrist mounting angles. The discrete data are fitted to obtain a continuous model (see Supplementary Materials). The yellow line indicates the thrust-drag equilibrium state among various configurations of flapping frequencies and angles of attack. (b) The cycle-averaged lift of the thrust-drag equilibrium state of different wrist mounting angles versus angles of attack. The black dashed line indicates the lift-weight equilibrium state. The wrist mounting angle affects the angle of attack of the equilibrium flight state. (c) The roll moment of different flapping frequencies in one wingbeat cycle, with the shaded region indicating the upstroke. The roll moment during upstroke basically remains the same, while the peak value of the roll moment in the downstroke phase becomes larger as the frequency increases. (d) The cycle-averaged roll moment of different angles of attack correlates linearly with the flapping frequency.

averaged lift would be equal to the weight of the RoboFalcon (600g). The equilibrium state is expressed as:

$$\begin{cases} \overline{T}_{net} = 0 \\ \overline{L} = W \end{cases} \quad (2)$$

For a given airspeed value, the cycle-averaged lift and net thrust are the functions of the angle of attack and the flapping frequency. Since the data of various flight states measured from the wind tunnel are discrete, it is hard to find the lift-weight and thrust-drag equilibrium state directly. To overcome that, we fit the cycle-averaged lift and cycle-averaged net thrust data with quadratic polynomial (see Section VII). In Fig. 7(a), the yellow line indicates the thrust-drag equilibrium state. For the net thrust graph, this is the level straight line past the zero point, while for the lift graph, this curve indicates the lift of the thrust-drag equilibrium state versus the angle of attack.

We compare the lift curves of the thrust-drag equilibrium state of the four wrist mounting angles in Fig. 7(b) with the lift-weight equilibrium state indicated by the black dashed line. The intersection of the black dashed line and the lift curves indicates the lift-weight and thrust-drag equilibrium states of the steady level flight.

The results indicate that a smaller θ allows the vehicle to fly at a larger trimmed angle of attack. This could be because a smaller θ involves a stronger downstroke twist of the hand-wing, and thus the vehicle requires a larger trimmed angle of attack to compensate for the pitch-down (wrist hyperpronating) of the wing to keep the wing's local angle of attack in the right range. Also, the mounting angle θ affects the available lift (the region of lift above 600g in Fig. 7(a) and thrust (the region of thrust above the equilibrium line in Fig. 7(a) at this airspeed. For our trials [Fig. 7(a)], the available lift increases with θ, while the available thrust maximizes at θ=20°. This suggests that birds and bats could use wrist supination-pronation movements to adjust the angle of attack of the steady level flight state and modulate the lift and thrust magnitude.

Then, we measure the roll moment of a 25° wrist mounting angle wing configuration at an airspeed of 8m/s under various flapping frequencies and angles of attack (0° to 12°). The average lift increases with the flapping frequency for all wing configurations, resembling previous studies on bendable or foldable flapping wing aerodynamics [41], [54], [55]. Since the roll moment of the RoboFalcon is achieved by the bilateral lift differential generated by the bilateral asymmetric downstroke, the correlation between lift and flapping frequency is also shown in the roll moment. Fig. 7(c) presents the roll moment of the morphing-coupled flapping that applies the bilateral asymmetric downstroke in a wingbeat cycle (α=12°). The cycle-averaged rolling moments of different angles of attack versus flapping frequency are illustrated in Fig. 7(d). In general, the roll moment remains at zero during the upstroke and only magnifies during the asymmetric downstroke. For the higher flapping frequency, the larger the peak value of the roll moment (negative values correspond to left roll). This means that the aerodynamic forces during downstroke dominate the creation of the roll moment, which is similar to the results of the study on birds [30].

## VII. Materials and Methods

### A. Bio-inspired morphing wing design and fabrication

The wing skeleton mechanism is a one-DOF sophisticated multi-link mechanism consisting of two sets of four-bar linkage mechanisms. We determine the lengths of the humeral spar, radial spar, and hand wing, using the vector closed-loop equation to solve the lengths of the link rods and levers by giving the angles of the shoulder, elbow, and wrist at the fully expanded and completely tucked states of the wing skeleton (see Supplementary Text for details). The lengths of the spars and hand wings are determined concerning the statistical bird (bat) wing data [58], [59].

We mainly construct the wing skeleton from carbon fiber composite boards. The link rods, levers, and fingers are directly cut from the 3k carbon fiber board (thicknesses of 2 and 1.5 mm) with a CNC milling machine (JINGDIAO Carver S400). The humeral and radial spars are made from the same carbon fiber board parts that are cut into a specific shape to form a 3D puzzle structure to constitute the hinge knuckles at two ends of the spars. The leading-edge support parts, wrist hinge knuckles, and other accessories, e.g., ball joint and servo mounting base, and are 3D printed (Raise3D Pro2 Plus) utilizing the Polylactic (PLA) and Polycarbonate (PC) materials. We employ the KINGMAX KM0940 digital servo (stall torque of 4.3kg·cm) for the wing morphing actuating, while for the wing membrane, the ripstop polyester fabric (210T), which is unstretchable and airtight. The wing membrane is only glued to the fingers, main spars, and leading-edge support parts of the skeleton to ensure no stretching during the skeleton folding process.

### B. Rolling & gliding controller

The rolling and gliding controller is constructed to generate a one-shot roll command for the rolling maneuver to test and control the wing stop position for the gliding state. The controller is equipped with an STM32F103C8T6 microprocessor running at 72MHz that contains a 6-channel PWM input to receive the operation command from the autopilot and a 3-channel PWM output to control the motor speed and the two servos for wing morphing actuation [Fig. 6(B)]. An OH3144 unipolar Hall-effect switch is connected to the microprocessor via an external interrupt.

When wings flap down to the level position, the magnet embedded in the main gear triggers the Hall switch to send an interrupt signal to the rolling and gliding controller. During flapping, the controller measures the time interval of two adjacent interrupt signals and calculates the flapping frequency.

When the rolling and gliding controller receives a predefined one-shot roll command, it takes over the roll control authority from the autopilot and waits for the Hall switch signal. Once the controller receives the signal, it delays half a wingbeat cycle and then outputs a left-wing morphing command with a duration of three-quarters of a wingbeat cycle to cover the next downstroke so that we can get a single bilateral asymmetric downstroke for the rolling maneuver test.

When the throttle is lower than a certain threshold, the controller reduces the motor speed and stops the motor after receiving the external interrupt signal. This affords to stop the wings right at the level position, and the glide-lock limits the gear from reversing to handle the aerodynamic load during gliding [Fig. 6(a)].

### C. Rolling maneuver test set up

We perform five rolling maneuver tests for recording the open-loop dynamic response of a single bilateral asymmetric downstroke. The tests are performed within a hangar to avoid wind interference, and the vehicle is tethered with a thin nylon string (diameter 0.55mm) for protecting itself from hitting ground and walls. The nylon string is slack during the flight and is light enough, presenting negligible additional weight and drag. The RoboFalcon is launched at the gliding state by throwing it. Immediately afterward, the motor starts and the vehicle goes into a closed-loop controlled stabilized flapping flight. Then, the operator sends a predefined roll command to the autopilot, enabling the rolling and gliding controller to take over the roll control and perform the open-loop roll maneuver initiated with the bilateral asymmetric downstroke.

The autopilot, Pixmini, is used for logging linear acceleration, angular velocity, attitude angle, airspeed, power consumption, and the control inputs. The airspeed sensor (sensirion sdp31) and the control inputs are sampled at 100Hz and 50Hz, respectively, while the remaining data at 200Hz. We calculate the mean and standard error of the attitude angle data for the five rolling maneuver flights, illustrated in Fig. 6(d). All data timings are normalized by the wingbeat cycle and aligned to the wingbeat that produced the asymmetric downstroke. The initial yaw angle is aligned to zero.

### D. Wind tunnel measurements

We use a low-speed low turbulence intensity wind tunnel housed at Northwestern Polytechnical University. The airspeed is provided by the wind tunnel precisely, and a tilting platform accurately controls the angle of attack. We use a PWM signal generator to control the throttle value for the flapping frequency adjustment, and the accurate frequency value is determined by recording the time interval of two adjacent signals from the onboard Hall-effect switch. We use a 6-axis load cell (ME K6D40 force/torque sensor) for the aerodynamic forces measurements, with the sample rate set to 1000Hz. The RoboFalcon is mounted on top of the load cell, while the latter is fixed with the tilting platform. With this installation, the gravity of RoboFalcon needs to be offset at every tested angle of attack. Since the lift and thrust are acquired in the coordinate system of the load cell, it is necessary to convert the acquired forces into the wind axis coordinate system for level steady flight analysis. The coordinate system conversion is as follows

$$\begin{bmatrix} L \\ T_{net} \end{bmatrix} = \begin{bmatrix} \sin\alpha & \cos\alpha \\ \cos\alpha & -\sin\alpha \end{bmatrix} \left( \begin{bmatrix} F_x \\ F_z \end{bmatrix} + G \begin{bmatrix} \sin\alpha \\ \cos\alpha \end{bmatrix} \right) \quad (3)$$

where the $L$ is the lift, $T_{net}$ the net thrust, $\alpha$ the angle of attack, $G$ the gravity of RoboFalcon, $F_x$ the force measured in the x-axis of the load cell, and $F_z$ the force measured in the z-axis of the load cell.

For each wing configuration measurement, the angle of attack ranges from 0 to 12 degrees with a 2-degree increment, while the throttle value increases with a 5% increment to cover the flapping frequency approximately from 2 to 4 Hz. Each state combining an angle of attack and a flapping frequency is measured for more than five wingbeat cycles. Each state's lift and net thrust data is cycle-averaged, and the maximum RMSE for different cycles is lower than 36.96g and 8.35g, respectively. We fit the data utilizing a quadratic polynomial to obtain a continuous model and determine the equilibrium point of each wing configuration in Fig. 7(a) and 7(b). The fitted results are presented in Supplementary Materials.

For the roll moment measurement, the angle of attack ranges from 0 to 12 degrees with a 4-degree increment, and the flapping frequency is roughly limited within the range of 1.4~2.6 Hz to protect the load cell. The curve of roll moment versus normalized time in Fig. 7(c) is filtered with the 5th order Butterworth low-pass filter (cut-off frequency of 12Hz), and the maximum RMSE of the cycle-averaged roll moment in Fig. 7(d) is lower than 0.007N·m.

## VIII. DISCUSSION

Inspired by the wingbeat kinematics of birds and bats during level flight, we develop a flapping wing aerial vehicle (FWAV), entitled RoboFalcon, equipped with the novel Conical Rocker Mechanism (CRM) to achieve this morphing-coupled flapping pattern. The CRM provides the coupling motions required for wing flapping and morphing and makes this platform achieve the wingbeat kinematics where the wings are fully extended during the downstroke and tucked during the upstroke. The untethered outdoor flight tests demonstrate the appealing flight ability achieved by our RoboFalcon and highlight the effectiveness of the proposed wingbeat kinematics.

Additionally, our design introduces a mechanical decoupler to allow for the independent control of wing morphing. The actuation source of the wing morphing alternates during flapping between the BLDC motor and servo, enabling an adjustable flapping pattern for each side and managing a rolling maneuver. To validate the agility improvement of our roll control strategy, we perform several rolling maneuver experiments employing the rolling and gliding controller. The results highlight that the bilateral asymmetric downstroke enables RoboFalcon to perform a 90-degree roll maneuver within only 2.5 wingbeat cycles, which is considered a highly appealing performance among nature flyers and bio-inspired FWAVs. The wind tunnel data indicate that the roll moment generated by the bilateral asymmetric downstroke increases with the flapping frequency, similar to the increase in the lift with frequency, which has been validated in previous studies of bendable or foldable flapping wings [41], [54], [55]. However, it is important to note that the increase in roll moment for a bird- or bat-inspired FWAV has not been reported yet. The frequency-regulated roll moment affords RoboFalcon a high roll control efficiency. During RoboFalcon's flight, the downstroke dominates in creating the roll moment during one wingbeat cycle, consistent with the previous study on birds [30].

Furthermore, our designed bio-inspired morphing wing applies a wrist mounting angle that affords to fine-tune the passive torsion of the hand wing during the downstroke. The wind tunnel data indicate that the angle of attack of the equilibrium state decrease with the increase of the wrist mounting angle. This result suggests that the wrist supination-pronation movements of birds and bats could be used to trim the angle of attack in a steady level flight. Adjustment of the wrist mounting angle provides an engineering solution to optimize the lift and thrust combination of the morphing-coupled flapping wing.

This work results in a novel actuation strategy for a flapping wing vehicle to couple the wing-morphing motion with the wingbeat motion. It introduces a new rolling attitude control of the bird- and bat-inspired FWAV, simplifying roll control with complex flapping kinematics and reducing the burden on the wing morphing actuator. Additionally, our setup allows exploiting less powerful and lighter-weighted actuators (like typical servo motor) and ultimately meeting the weight constraints during flight.

The wingbeat kinematics of RoboFalcon is very similar to that of birds, enabling a high level of bionic performance. Therefore, our platform is suitable for studying birds' and bats' aerodynamic and kinetic mechanisms during steady-level flight and maneuvers with bilateral asymmetric wingbeat patterns. The bat-style wing of RoboFalcon can accommodate the flight requirements well despite the membrane wrinkles when the wing is tucked. Our developed wing adopts several bionic features, i.e., 3D printed rigid leading-edge support and the raised elbow joint to mimic the function of the propatagium [43]-[45], which provides a design reference for the bio-inspired morphing wings based on vertebrate-wing anatomy to increase the camber of their inner section (arm wing). The final platform is agile and can fly quite well. Based on wind tunnel data, the RoboFalcon has considerable available lift beyond the equilibrium point and thus has significant load capacity to carry larger capacity batteries and onboard equipment for more extended missions. The RoboFalcon is also more stealthy due to the similarity of its wingbeat pattern to that of a bird.

It should be noted that, in this study, the wing based on the unstretchable membrane may generate sufficient lift for flight, but eventually, the efficiency is reduced due to the additional drag caused by the wrinkles. Since even in nature, birds fly more efficiently than bats [15], future work shall investigate a bird-style feathered wing for this platform, to allow an enhanced aerodynamic profile when the wings are tucked [34]. This work mainly focuses on providing an engineering solution to achieve the morphing-coupled flapping motion and validating the agility benefits from the bilateral asymmetric downstroke without delving into the energy efficiency advantage of this wingbeat pattern. Therefore future works shall also involve more detailed and precise experimental measurements to analyze the efficiency advantages of this wingbeat pattern after equipping some sort of single-degree-of-freedom feathered morphing wings. Also, this platform can only achieve the wingbeat pattern of birds and bats in level flight, but not the completely different pattern in the take-off and perching states used by these flying vertebrates. Considering that the CRM within RoboFalcon has the potential ability to adjust the flapping amplitude, stroke plane, dihedral angle, and sweep angle, RoboFalcon can be expected to couple these parameters with flapping by deploying appropriate actuation strategies to achieve the wingbeat patterns of birds or bats during take-off and perching, such as stroke plane tilting and amplitude modulation [29], [56], [57]. The latter shall drive research towards developing a flyable bio-inspired aerial robot that can mimic most of the wingbeat kinematics of birds and bats with complex degrees of freedom.

## APPENDIX

### Multi-link mechanism design of the morphing wing skeleton

To design the morphing wing skeleton, the lengths of the humerus ($l_h$), radius ($l_r$), and manus ($l_m$) need to be given. It is also necessary to determine the angle of the shoulder joint ($\theta_s$), elbow joint ($\theta_e$), wrist joint ($\theta_w$), and the displacement of the OS (output slider, $x_A$) when the skeleton is fully extended and tucked (subscripted by $e$ and $t$). Given these parameters, as shown in Fig. 8 and Table I, the vector closed-loop equation can be used to solve for the length of the remaining linkages.

Referring to Fig. 8(c), the vector closed-loop equations for the one slider-rocker mechanism and the two four-bar linkage mechanisms in the wing skeleton structure can be expressed as

$$\overrightarrow{OA} = \overrightarrow{OC} + \overrightarrow{CA} \tag{4}$$

$$\overrightarrow{CB} + \overrightarrow{BE} + \overrightarrow{ED} + \overrightarrow{DC} = 0 \tag{5}$$

$$\overrightarrow{EF} + \overrightarrow{FH} + \overrightarrow{HG} + \overrightarrow{GE} = 0 \cdot \tag{6}$$

The orthogonal decomposition of the above equation along with the xy direction yields

$$\begin{cases} x_A = c\cos\theta_1 + (b+a)\cos\theta_2 \\ 0 = c\sin\theta_1 - (b+a)\sin\theta_2 \end{cases} \tag{7}$$

$$\begin{cases} b\cos\theta_2 + e\cos\theta_3 - i\cos\theta_4 - d\cos\theta_1 = 0 \\ -b\sin\theta_2 + e\sin\theta_3 + i\sin\theta_4 - d\sin\theta_1 = 0 \end{cases} \tag{8}$$

$$\begin{cases} f\cos\theta_3 - j\cos\theta_5 - h\cos\theta_6 + (g+i)\cos\theta_4 = 0 \\ f\sin\theta_3 + j\sin\theta_5 - h\sin\theta_6 - (g+i)\sin\theta_4 = 0 \end{cases} \cdot \tag{9}$$

The variables in Fig. 8(c) are related to the given parameters in Fig. 8, (a) and (b), as follows

$$\begin{cases} \theta_s = \theta_1 \\ \theta_e = \theta_1 + \theta_4 \\ \theta_w = \theta_4 + \theta_6 \\ l_h = c + d \\ l_r = g \end{cases} \cdot \tag{10}$$

Substituting (10) into (7), (8), (9), we obtain

$$\begin{cases} x_A = (l_h - d)\cos\theta_s + (b+a)\cos\theta_2 \\ 0 = (l_h - d)\sin\theta_s - (b+a)\sin\theta_2 \end{cases} \tag{11}$$

$$\begin{cases} b\cos\theta_2 + e\cos\theta_3 - i\cos(\theta_e - \theta_s) - d\cos\theta_s = 0 \\ -b\sin\theta_2 + e\sin\theta_3 + i\sin(\theta_e - \theta_s) - d\sin\theta_s = 0 \end{cases} \tag{12}$$

$$\begin{cases} f\cos\theta_3 - j\cos\theta_5 - h\cos(\theta_w - \theta_e + \theta_s) + (l_r + i)\cos(\theta_e - \theta_s) = 0 \\ f\sin\theta_3 + j\sin\theta_5 - h\sin(\theta_w - \theta_e + \theta_s) - (l_r + i)\sin(\theta_e - \theta_s) = 0 \end{cases} \tag{13}$$

By using the trigonometric relationship and substituting among (11), (12), (13) to eliminate the variable $\theta_2$, $\theta_3$, $\theta_5$, we obtain

$$(x_A - (l_h - d)\cos\theta_s)^2 + (l_h - d)^2\sin^2\theta_s = (b+a)^2 \tag{14}$$

$$(d\cos\theta_s + i\cos(\theta_e - \theta_s) - b\frac{x_A - (l_h - d)\cos\theta_s}{b + a})^2 +$$
$$(d\sin\theta_s - i\sin(\theta_e - \theta_s) + b\frac{(l_h - d)\sin\theta_s}{b + a})^2 = e^2 \quad (15)$$

$$(\frac{f}{e}(d\cos\theta_s + i\cos(\theta_e - \theta_s) - b\frac{x_A - (l_h - d)\cos\theta_s}{b + a}) -$$
$$h\cos(\theta_w - \theta_e + \theta_s) + (l_r + i)\cos(\theta_e - \theta_s))^2 +$$
$$(\frac{f}{e}(d\sin\theta_s - i\sin(\theta_e - \theta_s) + b\frac{(l_h - d)\sin\theta_s}{b + a}) -$$
$$h\sin(\theta_w - \theta_e + \theta_s) - (l_r + i)\sin(\theta_e - \theta_s))^2 = j^2 \quad (16)$$

Since $b$ and $f$ are already given in Table I, the above three equations can be used to construct two sets of equations for the extended and tucked state, respectively, via the parameters given in Table I to derive the remaining six unknown variables. The remaining parameters of each linkage length of the wing skeleton are also shown in Table I.

| Given parameter | | | Derived linkage length (mm) | |
|---|---|---|---|---|
| linkage length (mm) | $l_h$ | 110 | $a$ | 15.3646 |
| | $l_r$ | 180 | $c$ | 33.5769 |
| | $l_w$ | 370 | $d$ | 76.4231 |
| | $b$ | 20 | $e$ | 73.6915 |
| | $f$ | 30 | $g$ | 180 |
| extended state | $\theta_{se}$ | 51° | $h$ | 20.1844 |
| | $\theta_{ee}$ | 110° | $i$ | 16.9713 |
| | $\theta_{we}$ | 147° | $j$ | 202.8699 |
| | $x_{Ae}$ | 45mm | | |
| tucked state | $\theta_{st}$ | 20° | | |
| | $\theta_{et}$ | 41° | | |
| | $\theta_{wt}$ | 35° | | |
| | $x_{At}$ | 65mm | | |

The given partial linkage lengths and the extended and tucked state parameters are shown at the left as the design constraints. Other linkage lengths derived from kinematic relations are shown at right.

FITTING RESULTS OF CYCLE-AVERAGED LIFT AND NET THRUST

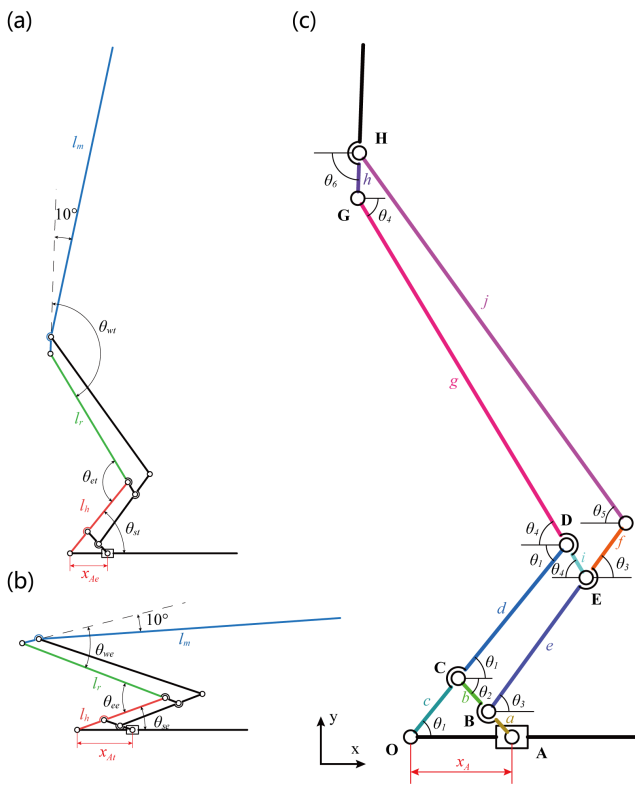

Fig. 8. Kinematic diagram of the morphing wing mechanism. The lengths of the humerus, radius, and manus are referenced to the forelimb structure of flying vertebrates while meeting the wing span constraints, which are the given design parameters. (a) and (b) show the fully extended and tucked states of the wing skeleton, respectively, and the angles of the shoulder, elbow, and wrist joints for these two states are given in the diagrams as the design constraints for determining the remaining linkage length. (c) Reference diagram for mechanism kinematic analysis.

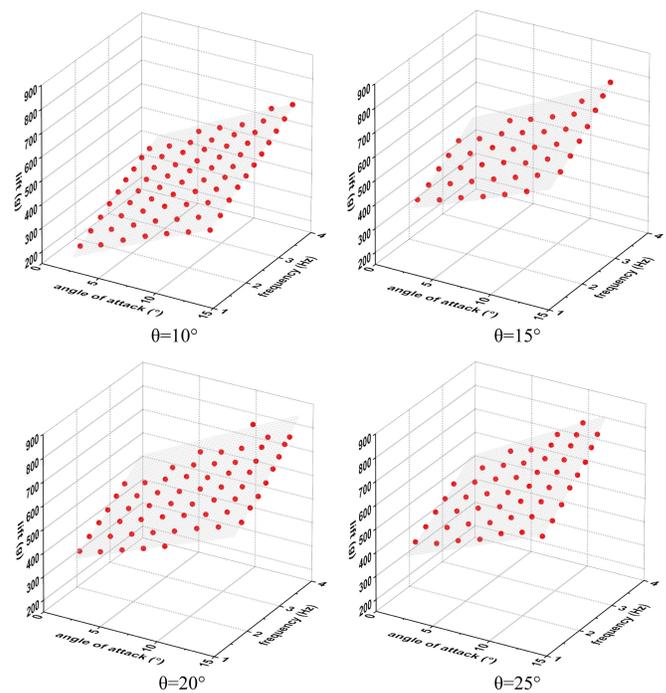

Fig. 10. Fitting surfaces of the cycle-averaged lift versus angle of attack and flapping frequency for different wrist mounting angles.

TABLE II
FITTING RESULTS OF CYCLE-AVERAGED LIFT AND NET THRUST

| | | Cycle-averaged lift | | | | Cycle-averaged net thrust | | | |
|---|---|---|---|---|---|---|---|---|---|
| Model | | $L = z_0 + a\alpha + bF + c\alpha^2 + dF^2 + f\alpha F$ | | | | $T = z_0 + a\alpha + bF + c\alpha^2 + dF^2 + f\alpha F$ | | | |
| Wrist mounting angle $\theta$ (°) | | 10 | 15 | 20 | 25 | 10 | 15 | 20 | 25 |
| Parameters | $z_0$ | 9.940 | 155.857 | 191.549 | 202.331 | -127.799 | -63.082 | -80.520 | -67.745 |
| | a | 22.879 | 12.322 | 14.278 | 13.544 | 1.402 | -0.231 | 2.939 | 2.594 |
| | b | 97.421 | 29.811 | 11.376 | 10.631 | 18.077 | -17.318 | -4.745 | -11.970 |
| | c | -0.095 | 0.092 | -0.029 | -0.038 | -0.222 | -0.213 | -0.417 | -0.444 |
| | d | -0.664 | 12.722 | 17.959 | 18.423 | 7.276 | 12.767 | 10.466 | 10.952 |
| | f | 0.576 | 2.769 | 2.656 | 3.454 | -0.453 | -0.6618 | -0.939 | -1.184 |
| R value | | 0.99969 | 0.99948 | 0.99936 | 0.99864 | 0.99952 | 0.99950 | 0.99825 | 0.99851 |
| RSME (g) | | 3.958 | 3.015 | 4.808 | 6.631 | 1.270 | 1.016 | 2.127 | 1.960 |
| Number of point | | 70 | 42 | 59 | 50 | 70 | 42 | 59 | 50 |

For the model expressions in the table, $L$ is the cycle-averaged lift, $T$ the cycle-averaged net thrust, $\alpha$ the angle of attack, and $F$ the flapping frequency.

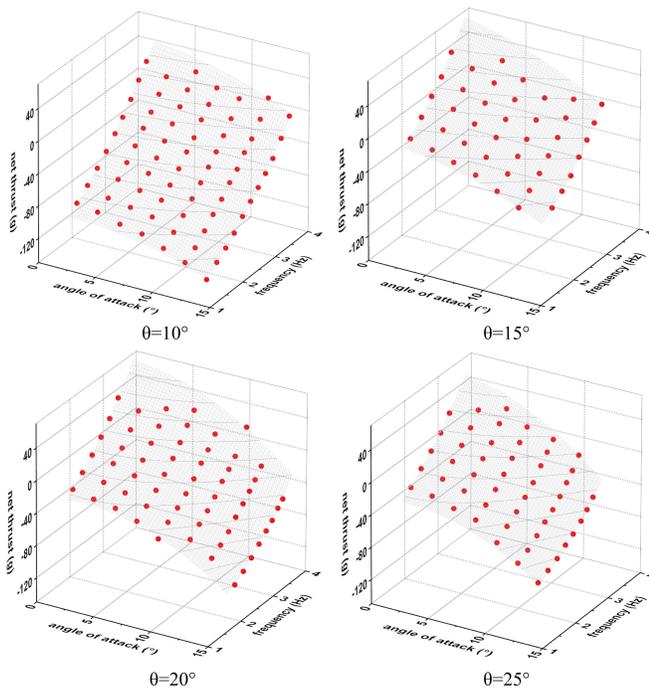

Fig. 10. Fitting surfaces of the cycle-averaged net thrust versus angle of attack and flapping frequency for different wrist mounting angles.